\documentclass[letterpaper, 10 pt, conference]{ieeeconf}  

\IEEEoverridecommandlockouts                              

\usepackage{amsmath} 
\usepackage{amssymb}  

\usepackage{graphicx}
\usepackage{capt-of} 
\usepackage{overpic} 
\usepackage[export]{adjustbox} 
\usepackage{textcomp} 
\usepackage{tabularx}
\usepackage{multirow}
\usepackage{booktabs}
\usepackage{arydshln}
\usepackage{adjustbox} 
\usepackage{pgfplots} 
\usepackage{units}
\usepackage[implicit=false]{hyperref}

\usepackage[caption=false, font=footnotesize]{subfig} 

\newcommand\copyrighttext{\footnotesize \textcopyright~2022 IEEE. Personal use of this material is permitted.  Permission from IEEE must be obtained for all other uses, in any current or future media, including reprinting/republishing this material for advertising or promotional purposes, creating new collective works, for resale or redistribution to servers or lists, or reuse of any copyrighted component of this work in other works.
}%

\newcommand\copyrightnotice{%
	\begin{tikzpicture}[remember picture,overlay]
	\node[anchor=south,xshift=0pt,yshift=14pt] at (current page.south) {\fbox{\parbox{\dimexpr\textwidth-\fboxsep-\fboxrule\relax}{\copyrighttext}}};
	\end{tikzpicture}%
}

\title{\LARGE \bf
CRAT-Pred: Vehicle Trajectory Prediction with Crystal Graph Convolutional Neural Networks and Multi-Head Self-Attention
}

\author{Julian Schmidt$^{1, 2}$, Julian Jordan$^{1}$, Franz Gritschneder$^{1}$ and Klaus Dietmayer$^{2}$
\thanks{\scriptsize $^{1}$Mercedes-Benz AG, R\&D, Stuttgart, Germany \newline \{ {\tt\scriptsize julian.sj.schmidt, julian.jordan, \newline franz.gritschneder} \} { \tt\scriptsize @mercedes-benz.com }
        }%
\thanks{\scriptsize $^{2}$Ulm University, Institute of Measurement, Control and Microtechnology, Ulm, Germany
        \{ {\tt\scriptsize Klaus.Dietmayer} \} { \tt\scriptsize @uni-ulm.de}}%
\thanks{\scriptsize$^{3}$Source code: \url{https://github.com/schmidt-ju/crat-pred}}%
}

\begin{document}

\maketitle
\thispagestyle{empty}
\pagestyle{empty}

\begin{abstract}
Predicting the motion of surrounding vehicles is essential for autonomous vehicles, as it governs their own motion plan.
Current state-of-the-art vehicle prediction models heavily rely on map information.
In reality, however, this information is not always available.
We therefore propose CRAT-Pred, a multi-modal and non-rasterization-based trajectory prediction model, specifically designed to effectively model social interactions between vehicles, without relying on map information.
CRAT-Pred applies a graph convolution method originating from the field of material science to vehicle prediction, allowing to efficiently leverage edge features, and combines it with multi-head self-attention.
Compared to other map-free approaches, the model achieves state-of-the-art performance with a significantly lower number of model parameters.
In addition to that, we quantitatively show that the self-attention mechanism is able to learn social interactions between vehicles, with the weights representing a measurable interaction score.
The source code is publicly available$^3$.
\end{abstract}

\section{Introduction}
\copyrightnotice 
In order to make autonomous vehicles safer than human drivers, there is a strong need to predict the future motion of vehicles participating in traffic.
Solving this task in an appropriate manner requires methods that consider the social interactions between these participating vehicles.
This can be illustrated by a simple thought experiment.
Imagine two vehicles driving behind each other on a straight road.
The future trajectory of the trailing vehicle is highly dependent on the current action of the leading vehicle.
Neglecting such strong social interactions leads to inappropriate results for safe autonomous driving tasks.

With regard to this challenge, particularly machine learning-based prediction models have shown strong performance in the past.
While earlier works relied on hand-crafted features to represent social interactions, latest work proposes models that are based on Convolutional Neural Networks (CNN) (e.g., \cite{Hong2019}), Graph Neural Networks (GNN) (e.g., \cite{Li2020a_ARXIV}) or the attention mechanism (e.g., \cite{Liang2020}).
A majority of these models for vehicle prediction have one thing in common:
Their architectural design and the corresponding training process is designed for the incorporation of map information, most commonly originating from an underlying High Definition (HD)-map.
In reality, however, map information is not always existent, up-to-date or usable due to failed localization.
This illustrates the need for high performant, map-free prediction models.
Although this also applies to the prediction of pedestrian motion, recent work \cite{Schoeller2020} has shown that even sophisticated neural network-based models cannot significantly outperform simple approaches, such as the constant velocity model.
For vehicles, the underlying kinematic model restricts their motion and the clearly defined traffic rules provide certain constraints to their interactions.
These two aspects of vehicle prediction, together with other, yet to be investigated effects, for now seem essential for learning the task with a machine learning-based approach.

In this work we propose CRAT-Pred \textit{(Crystal Attention Prediction)}, an interaction-aware and multi-modal trajectory prediction model for vehicles that does not rely on map information.
CRAT-Pred is specifically designed to make the best possible use of the information provided by social interactions.
For this purpose it uses the mechanisms of crystal graph convolutional neural networks \cite{Xie2018} that originated in the field of material science.
The multi-head self-attention mechanism, directly prior to the trajectory decoder, offers an additional way to learn social interactions in an interpretable way.

In summary, our main contributions are:
\begin{itemize}
	\item We propose a novel map-free trajectory prediction model for vehicles. To the best knowledge of the authors, this model is a first-time combination of crystal graph convolutional neural networks and multi-head self-attention.
	\item We conduct extensive experiments on the publicly available Argoverse Motion Forecasting Dataset \cite{Chang2019} and prove state-of-the-art performance for map-free prediction. In contrast to other approaches with similar performance, the model requires significantly less model parameters.
	\item We furthermore quantitatively show what has only been assumed so far: The self-attention mechanism is able to learn social interactions between vehicles.
\end{itemize}

\begin{figure*}[thpb]
	\centering
	\includegraphics[width=\textwidth, trim=0cm 0cm 0cm 0cm, clip]{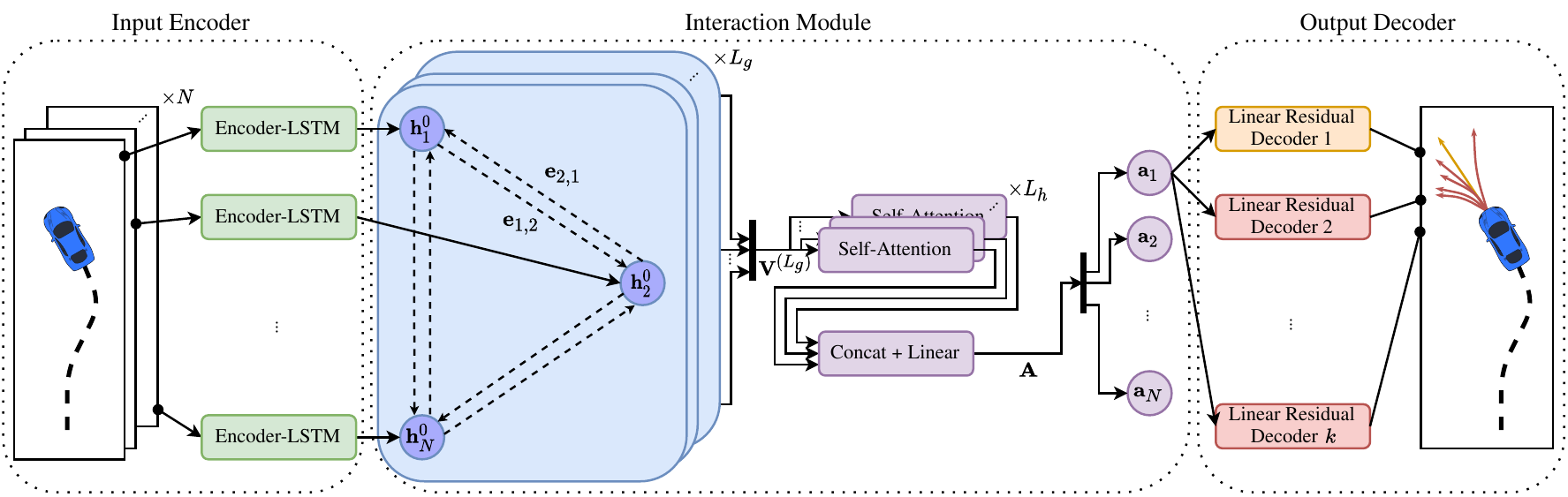}
	\vspace{-0.6cm}
	\caption{Overview of CRAT-Pred: Each actor is encoded by an LSTM (green) that shares weights between all actors. Interactions are modeled by a GNN (blue) and a multi-head self-attention mechanism (purple). Trajectory prediction is done by multiple parallel linear residual layers (orange and red), with the first one (orange) always corresponding to the most probable trajectory.}
	\label{fig:trajectory_prediction_model}
\end{figure*}

\section{Related Work}
Recent machine learning-based approaches to model social interactions for vehicle trajectory prediction can be clustered into two distinctive categories.

\textbf{Rasterization-based approaches}\quad rely on grid-based operations to model social interactions.
Deo et al. \cite{Deo2018} continue the work of the Social-LSTM \cite{Alahi2016} and encode vehicle dynamics via a Long Short-Term Memory (LSTM) followed by a convolutional social pooling mechanism for learning interdependencies between vehicles.
More advanced approaches rely on the generation of a top-down Bird's-Eye-View (BEV) image representation \cite{Hong2019, Chai2020, Djuric2020, Strohbeck2020}.
The BEV representation offers a flexible way to not only model social interactions, but also temporal vehicle information and even map information.
More specifically, different types of context information can be encoded in different channels of the image.
Common CNN backbones, such as ResNet \cite{Chai2020, Djuric2020} or MobileNetV2 \cite{Djuric2020, Strohbeck2020}, are then applied to extract mid-level features, which can then be interpreted by a decoder.
However, the flexibility comes with a drawback.
Generating BEV representations requires rasterization and thereby leads to an unavoidable loss of information \cite{Liang2020}.

\textbf{Node-based approaches}\quad on the other hand do not rely on rasterization, but model social interactions in a graph structure, with nodes typically representing the vehicles.
Our definition of node-based approaches includes GNN-based approaches \cite{Li2020a_ARXIV,Casas2020, Salzmann2020} that rely on message passing \cite{Gilmer2017} and attention-based approaches \cite{Liang2020, Tang2019, Messaoud2021, Mercat2020, Gao2020} that rely on the attention mechanism \cite{Vaswani2017}.
Due to the flexible node structure, there are also approaches combining both groups \cite{Khandelwal2020_ARXIV, Ma2019, Li2020b}.
The underlying graph is either fully-connected \cite{Liang2020, Casas2020, Messaoud2021, Mercat2020, Gao2020, Khandelwal2020_ARXIV} or dependent on the distance between the vehicles \cite{Li2020a_ARXIV, Salzmann2020, Tang2019, Ma2019, Li2020b}.
Analogously to some of the rasterization-based approaches, recurrent encoders are commonly used to get an encoding of the past vehicle states, thus making the temporal information available to the corresponding nodes.

Most of the above listed state-of-the-art approaches directly incorporate map information in a unified way into the graph structure.
For instance, VectorNet \cite{Gao2020} creates one global interaction graph with map information as additional nodes.
Another example is LaneGCN \cite{Liang2020}, which uses multiple attention mechanisms to pass information between vehicle nodes and the features of a lane graph.
Other recent approaches aggregate information of interacting vehicles directly through the road topology, resulting in a non-interpretable way of interaction modeling \cite{Kim2021_ARXIV, Zeng2021_ARXIV}.
Both ways to incorporate map information result in models that are tightly coupled, designed and trained for the incorporation of map information, which limits the application to regions where this information is available.

Mercat et al. \cite{Mercat2020} propose a prediction model that intentionally requires no map information and combines an LSTM encoder-decoder with multiple multi-head self-attention layers.
Experiments are only carried out for highway scenarios.
Messaoud et al. \cite{Messaoud2021} propose a model that combines grid-based trajectory encoding with LSTMs and multi-head self-attention, resulting in a hybrid rasterization-node-based approach that is limited to highway scenarios.
Our proposed model is related to the work of Mercat et al. \cite{Mercat2020} and Messaoud et al. \cite{Messaoud2021} and is therefore specifically designed for map-free prediction, thus relying on making the best possible use of the information provided by social interactions.
It shares the idea of using multi-head self-attention to model social interactions in an interpretable way.
In contrast to these approaches, however, it combines the powerful mechanisms of GNNs, including the otherwise rare usage of features for graph edges, and multi-head self-attention, resulting in an approach not limited to specific scenarios and not relying on rasterization.
We show this by benchmarking our model on a diverse, large-scale and publicly available dataset.

\textbf{Other recent approaches}\quad reuse algorithms from point cloud learning \cite{Ye2021} or utilize whole end-to-end transformer-inspired architectures \cite{Liu2021}.

\section{Problem Formulation}
Our formulation considers the trajectory prediction problem as the task to predict the future 2D coordinates of vehicles, given their past states and the past states of the surrounding vehicles.
In a scene with $N$ vehicles, the available information is
\begin{equation}
\mathbf{T}_\mathrm{hist} = \{ \boldsymbol{\tau}^t_i | i \in 1, \dots , N ; t \in -T_h + 1, \dots , 0 \} \text{,}
\end{equation}
where $T_h$ denotes the history horizon.
At each timestep $t$, vehicle $i$ is represented by the 2D coordinates $\boldsymbol{\tau}^t_i = (x^t_i, y^t_i)$.

Given this information, the trajectory prediction problem can be expressed as the task to predict
\begin{equation}
	\mathbf{T}_\mathrm{fut} = \{ \tilde{\boldsymbol{\tau}}^{l}_i | i \in 1, \dots , N ; l \in 1, \dots , T_f \} \text{,}
\end{equation}
with $T_f$ being the forecasting horizon and $\tilde{\boldsymbol{\tau}}^l_i = (x^l_i, y^l_i)$.

\section{Trajectory Prediction Model}
An architectural overview of our proposed trajectory prediction model with the name CRAT-Pred is given in Fig. \ref{fig:trajectory_prediction_model}.
The following sections provide an in-depth description of the individual components.

\subsection{Input Encoder}
Instead of using absolute 2D coordinates, our model operates on input data, which represents the past of each vehicle $i$ as a series of discrete displacements
\begin{equation}
	\mathbf{s}_i^t = (\Delta \boldsymbol{\tau}^{t}_i || b^{t}_i) \text{,}
\end{equation}
with $\Delta \boldsymbol{\tau}^{t}_i = \boldsymbol{\tau}^{t}_i - \boldsymbol{\tau}^{t-1}_i$.
Based on the input data representation of LaneGCN \cite{Liang2020}, we only consider vehicles that are observable at $t=0$ and handle vehicles that are not observed over the full history horizon $T_h$ by concatenating a binary flag $b_i^t$.
The flag indicates whether there was a displacement of vehicle $i$ observed at timestep $t$ ($b_i^t = 1$) or not ($\Delta \boldsymbol{\tau}^{t}_i = (0, 0)$ and $b^{t}_i = 0$).
For each vehicle $i$ and each timestep $t$, this results in a vector $\mathbf{s}_i^t$ of size $3$.

Given this information, a single LSTM
\begin{equation}
	\mathbf{h}_i^t = \mathrm{LSTM}(\mathbf{h}^{t-1}_i, \mathbf{s}^t_i, \mathbf{W}_\mathrm{enc}, \mathbf{b}_\mathrm{enc})
\end{equation}
with one layer and shared weights for all vehicles is then used to encode the temporal information of each vehicle in the scene.
The hidden state $\mathbf{h}_i^t$ is vector of size $128$.

\subsection{Interaction Module}
Subsequent to the encoding of the past state of each vehicle, we construct a bidirectional fully-connected interaction graph, with $\mathbf{v}_i^{(0)} = \mathbf{h}_i^0$ acting as the initial node features.
In addition to that, edge features are used:
The edge from node $i$ to node $j$ obtains the feature vector
\begin{equation}
	\mathbf{e}_{i,j} = \boldsymbol{\tau}^0_j - \boldsymbol{\tau}^0_i \text{,}
\end{equation}
which corresponds to the distance from vehicle $i$ to vehicle $j$ at $t=0$.

The graph convolution operator is then defined as the one used by crystal graph convolutional neural networks \cite{Xie2018}
\begin{multline}
	\mathbf{v}_i^{(g+1)} = \mathbf{v}_i^{(g)} + \\
	\sum_{j \in \mathcal{N}(i)}\sigma \left( \mathbf{z}_{i,j}^{(g)} \mathbf{W}_\mathrm{f}^{(g)} + \mathbf{b}_\mathrm{f}^{(g)} \right)
	\odot g \left( \mathbf{z}_{i,j}^{(g)} \mathbf{W}_\mathrm{s}^{(g)} + \mathbf{b}_\mathrm{s}^{(g)}  \right) \text{.}
\end{multline}
These were originally developed for the prediction of material properties and, to the best of our knowledge, we first-time apply them to the prediction of vehicles.
In contrast to many other graph convolution operators, it is designed for the incorporation of edge features, in our case allowing the network to additionally update the node features based on the distance between vehicles.
$g \in 0, \dots , L_g$ denotes the layer of the GNN, with $L_g$ corresponding to the total number of layers.
We use $L_g = 2$, with batch normalization and $\mathrm{ReLU}$ as non-linearity between the layers.
Deeper GNNs, where $L_g > 2$, are possible, if there is a need to model more complex interactions.
$\mathbf{z}_{i,j}^{(g)} = ( \mathbf{v}_i^{(g)} || \mathbf{v}_j^{(g)} ||\mathbf{e}_{i,j} )$ corresponds to the concatenation of the node features and the edge feature. $\sigma$ and $g$ are a sigmoid and softplus function.

After the GNN, each updated node feature $\mathbf{v}_i^{(L_g)}$ holds information about a vehicle and its social context, however, vehicles might still be required to pay attention to specific surrounding vehicles, depending on their past trajectory and current position.
In order to model this, we use a scaled dot-product multi-head self-attention layer \cite{Vaswani2017} and apply it to the updated node feature matrix $\mathbf{V}^{(L_g)}$, which contains the node features $\mathbf{v}_i^{(L_g)}$ as rows, resulting in a shape of $N \times 128$.
Each head $h \in 1,\dots, L_h$ is defined as
\begin{equation}
	\mathrm{head}_h = \mathrm{softmax} \left( \frac{\mathbf{V}^{(L_g)}_{Q_h} \mathbf{V}^{(L_g) T}_{K_h}}{\sqrt{d}}  \right) \mathbf{V}^{(L_g)}_{V_h} \text{.}
\end{equation}
$\mathbf{V}^{(L_g)}_{Q_h}$, $\mathbf{V}^{(L_g)}_{K_h}$ and $\mathbf{V}^{(L_g)}_{V_h}$ are head $h$s' linear projections of the node feature matrix $\mathbf{V}^{(L_g)}$ and $d$ is a normalization factor that corresponds to the embedding size of each head.
The result of the softmax-function is often referred to as the attention weight matrix, in this case having a shape of $N \times N$ and representing pairwise dependencies between vehicles.
We will later experimentally analyze the attention weights more in-depth.

Finally, the updated node feature matrix $\mathbf{A}$ is obtained by
\begin{equation}
	\mathbf{A} = (\mathrm{head}_1 || \dots || \mathrm{head}_{L_h}) \mathbf{W}_\mathrm{o} + 
	\begin{pmatrix}
	        \mathbf{b}_\mathrm{o}\\
			\vdots \\
			\mathbf{b}_\mathrm{o}
	\end{pmatrix}.
\end{equation}
We use $L_h = 4$ and $d= \frac{128}{L_h} = 32$.
This means that one row $\mathbf{a}_i$ of the feature matrix $\mathbf{A}$ is a vector of size $128$ and corresponds to the interaction-aware features of vehicle $i$.

\subsection{Output Decoder}
For trajectory prediction, we use a linear residual layer and apply a linear projection to it.
Instead of directly predicting the 2D coordinates in the global coordinate frame, the output decoder predicts the positional differences $\mathbf{o}^l_i = \tilde{\boldsymbol{\tau}}^l_i - \boldsymbol{\tau}^0_i$ of vehicle $i$ to its 2D coordinates at $t=0$.
Formally, the output decoder is defined as
\begin{equation}
	\boldsymbol{o}_{i} = (\mathrm{ReLU} (\mathcal{F}(\mathbf{a}_i, \{ \mathbf{W}_\mathrm{r}, \mathbf{b}_\mathrm{r} \}) + \mathbf{a}_i)) \mathbf{W}_\mathrm{dec} + \mathbf{b}_\mathrm{dec} \text{,}
\end{equation}
with
\begin{equation}
	\mathcal{F} = (\mathrm{ReLU} (\mathbf{a}_i \mathbf{W}_{\mathrm{r}, 2} + \mathbf{b}_{\mathrm{r}, 2})) \mathbf{W}_{\mathrm{r}, 1} + \mathbf{b}_{\mathrm{r}, 1} \text{.}
\end{equation}
Group norm is used for normalization.

Multi-modality is obtained by using $k$ of these decoders in parallel.
Further details are given in the next section.

\subsection{Training}
Current approaches for making multi-modal predictions use an additional classification layer to determine the probability of each individual mode.
This not only adds model complexity, but also turns the learning process into a multi-task problem, which can result in problems regarding loss balancing and convergence.
We claim that it is indispensable to identify a vehicle's most probable trajectory, but the probabilities of all other modes have a subordinate role.
It should be noted, however, that this is highly dependent on the subsequent planning algorithm.

Therefore, we obtain multi-modality by first training the full network end-to-end with only one output decoder, always resulting in the most optimal prediction.
The loss function used for this training step is smooth-L1 loss.
After convergence, we freeze the whole model and add $k - 1$ additional learnable output decoders to it.
These additional decoders are then trained with Winner-Takes-All (WTA) loss \cite{Guzmanrivera2012}.
In this specific case, WTA means that for each sequence only the weights of the decoder with the smallest smooth-L1 loss are optimized.

\section{Experiments}
The following sections describe the extensive evaluation of our model on the widely established Argoverse Motion Forecasting Dataset \cite{Chang2019}.
We prove state-of-the-art performance for map-free prediction.
In addition to that, we quantitatively show that the weights resulting from the multi-head self-attention layer are a superior indicator for the pairwise interaction of vehicles, compared to the Euclidean distance.

\subsection{Dataset}
The Argoverse dataset consists of $205{,}942$ train, $39{,}472$ validation and $78{,}143$ test sequences recorded in Miami and Pittsburgh, each containing trajectories of multiple vehicles sampled with \unit[$10$]{Hz}.
The goal is to predict the future trajectory ($3$ seconds) of one target vehicle, while taking the past trajectories ($2$ seconds) of all vehicles in a sequence into account.
Sequences in the train and validation set therefore have a length of $5$ seconds, while sequences in the test set only contain the first $2$ seconds of motion.
While the trajectory of the target vehicle is guaranteed to be observed over the full $5$ seconds, other vehicle trajectories are possibly only partially observed during this duration.
Due to the intended map-free design of our proposed model, we do not utilize the HD-maps provided by the dataset.

\subsection{Metrics}
For evaluation, we follow the previous works and adopt the minimum Average Displacement Error (minADE), the minimum Final Displacement Error (minFDE) and the Miss Rate (MR) for single- ($k=1$) and multi-modal ($k=6$) predictions.
minADE corresponds to the minimum average Euclidean error between the predicted trajectory and the ground-truth trajectory of the target vehicle, while considering the top $k$ predictions.
Analogously minFDE corresponds to the minimum Euclidean error between the predicted endpoint and the ground-truth endpoint.
MR is defined as the ratio of sequences where none of the predicted endpoints is closer than $2$ meters to the ground-truth endpoint.

\subsection{Implementation Details}
During preprocessing, coordinate transformation of each sequence into a local target vehicle coordinate frame is done.
This common preprocessing step is also performed by other approaches \cite{Liang2020, Ye2021} benchmarked on the Argoverse dataset.
Therefore, the coordinates in each sequence are transformed into a coordinate frame originated at the position of the target vehicle at $t=0$.
The orientation of the positive x-axis is given by the vector described by the difference between the position at $t=0$ and $t=-1$.

The model is trained for $72$ epochs using Adam optimizer \cite{Kingma2015} with a batch size of $32$ and a weight decay of $10^{-2}$.
The first $36$ epochs are used to train the full network end-to-end with only one output decoder.
After $32$ epochs the learning rate decays from $10^{-3}$ to $10^{-4}$.
The subsequent $36$ epochs are used to train the additional learnable output decoders, with the initially trained model weights frozen.
Learning rates and decay are applied in the same way.
Implementation is done with pytorch \cite{Paszke2019} and pytorch geometric \cite{Fey2019}.

\subsection{Quantitative Results}
Table \ref{tab:results_test_map_free} compares our model to the map-free baselines \cite{Chang2019} on the online evaluated Argoverse test set.
Our model outperforms all of them by a large margin.
Fig. \ref{fig:performance_comparison} visualizes the minADE of these map-free baselines and the minADE of state-of-the-art models that include map information.
Our model yields competitive results even in direct comparison to the state-of-the-art models, despite solely focusing on the social interactions and not including map information.

As already stated, most state-of-the-art models for vehicle trajectory prediction are specifically designed for the incorporation of map information and cannot be used and evaluated for map-free prediction.
Nevertheless, there are a few models that can be adapted and then trained for map-free prediction.
For a more in depth performance analysis, Table \ref{tab:results_val} compares our model with the results of other adapted current state-of-the-art models for map-free prediction on the Argoverse validation set.
In this case, our model manages to outperform all current state-of-the-art models in minADE$_{@k=1}$ and MR$_{@k=1}$, while requiring a significantly lower number of model parameters.
This is a strong indicator that our proposed model is a high performant and simultaneously efficient method for vehicle trajectory prediction.

\begin{table*}[thpb]
	\begin{minipage}{\columnwidth}
		\caption{Results on the Argoverse test set: Map-free}
		\label{tab:results_test_map_free}
		\setlength{\tabcolsep}{2.3pt}
		\centering
		\begin{tabularx}{\columnwidth}{Xlllllll}
			\toprule
			\multirow{2}{*}{Method}       &              \multicolumn{3}{c}{$k=1$}              &              \multicolumn{3}{c}{$k=6$}              &  \\
			                              & minADE          & minFDE          & MR              & minADE          & minFDE          & MR              &  \\ \midrule
			LSTM ED \cite{Chang2019}      & $2.15$          & $4.97$          & $0.75$          & -               & -               & -               &  \\
			LSTM ED-soc. \cite{Chang2019} & $2.15$          & $4.95$          & $0.75$          & -               & -               & -               &  \\
			NN \cite{Chang2019}           & $3.45$          & $7.88$          & $0.87$          & $1.71$          & $3.29$          & $0.54$          &  \\
			CVM \cite{Chang2019}          & $3.53$          & $7.89$          & $0.83$          & -               & -               & -               &  \\ \midrule
			Ours                          & $\mathbf{1.82}$ & $\mathbf{4.06}$ & $\mathbf{0.63}$ & $\mathbf{1.06}$ & $\mathbf{1.90}$ & $\mathbf{0.26}$ &  \\ \bottomrule
		\end{tabularx}
	\end{minipage}
	\hfill
	\begin{minipage}{\columnwidth}
		\vspace{0.18cm}
		\centering
		\begin{adjustbox}{clip, trim=0cm 0.1cm 0cm 0cm}
			\resizebox{\textwidth}{!}{\input{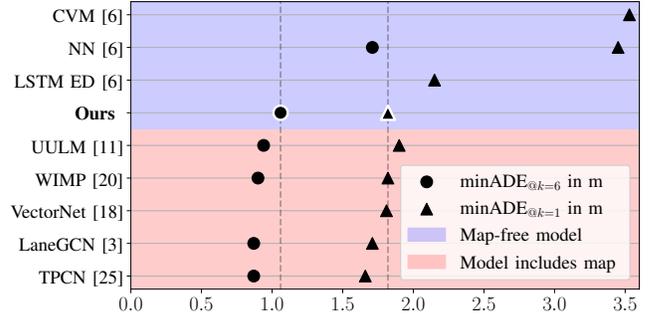}}
		\end{adjustbox}
		\vspace{-0.6cm}
		\captionof{figure}{minADE$_{@k=1}$ and minADE$_{@k=6}$ comparison of map-free models and models that include map information.}
		\label{fig:performance_comparison}
	\end{minipage}
\end{table*}
\begin{table*}[thpb]
	\caption{Results for map-free prediction on the Argoverse validation set}
	\label{tab:results_val}
	\setlength{\tabcolsep}{2.9pt}
	\centering
	\begin{tabularx}{1\textwidth}{Xlllllll}
		\toprule
		\multirow{2}{*}{Method}                                          & \multirow{2}{*}{Number of Model Parameters} &              \multicolumn{3}{c}{$k=1$}              &              \multicolumn{3}{c}{$k=6$}              \\
		                                                                 &                                             & minADE          & minFDE          & MR              & minADE          & minFDE          & MR              \\ \midrule
		TPCN \cite{Ye2021}                                               & -                                           & $1.42$          & $\mathbf{3.08}$ & $0.55$          & $0.82$          & $1.32$          & $\mathbf{0.15}$ \\
		LaneGCN \cite{Liang2020}                                         & $1{,}017{,}769$                             & $1.58$          & $3.61$          & -               & $\mathbf{0.79}$ & $\mathbf{1.29}$ & -               \\
		WIMP \cite{Khandelwal2020_ARXIV}                                 & $> 20{,}000{,}000$                          & $1.61$          & $5.05$          & -               & $0.86$          & $1.39$          & $0.16$          \\ \midrule
		Ours (LSTM + GNN + Lin. Residual)                                & $\mathbf{448{,}872}$                        & $1.44$          & $3.17$          & $0.54$          & $0.86$          & $1.47$          & $0.19$          \\
		Ours (LSTM + GNN + Multi-Head Self-Attention + Lin. Residual)    & $514{,}920$                                 & $\mathbf{1.41}$ & $3.10$          & $\mathbf{0.52}$ & $0.85$          & $1.44$          & $0.17$          \\ \bottomrule
	\end{tabularx}
\end{table*}

\subsection{Qualitative Results}
\begin{figure*}[thpb]
	\centering
	\subfloat{%
		\includegraphics[width=0.33\textwidth, trim=15cm 13cm 12.5cm 11.5cm, clip, frame]{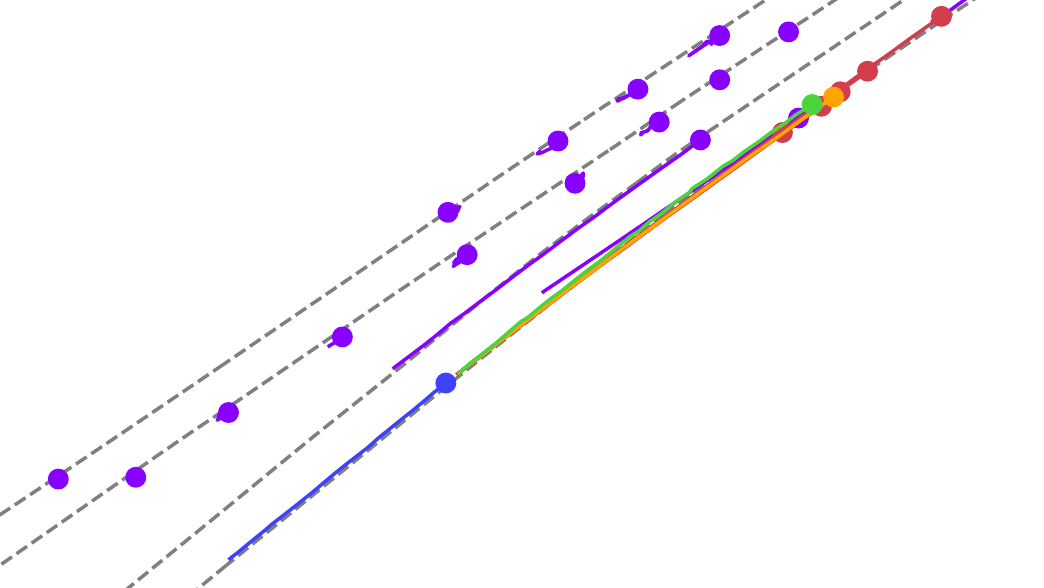}}
	\hfill
	\subfloat{%
		\includegraphics[width=0.33\textwidth, trim=14.5cm 11.5cm 13cm 13cm, clip, frame]{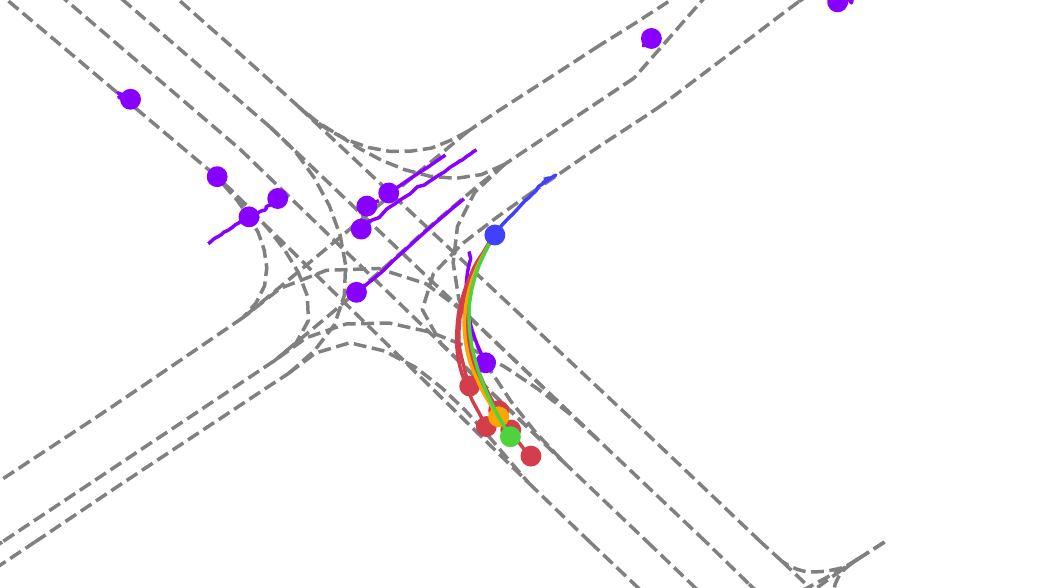}}
	\hfill
	\centering
	\subfloat{%
		\includegraphics[width=0.33\textwidth, trim=14cm 12.5cm 13.5cm 12cm, clip, frame]{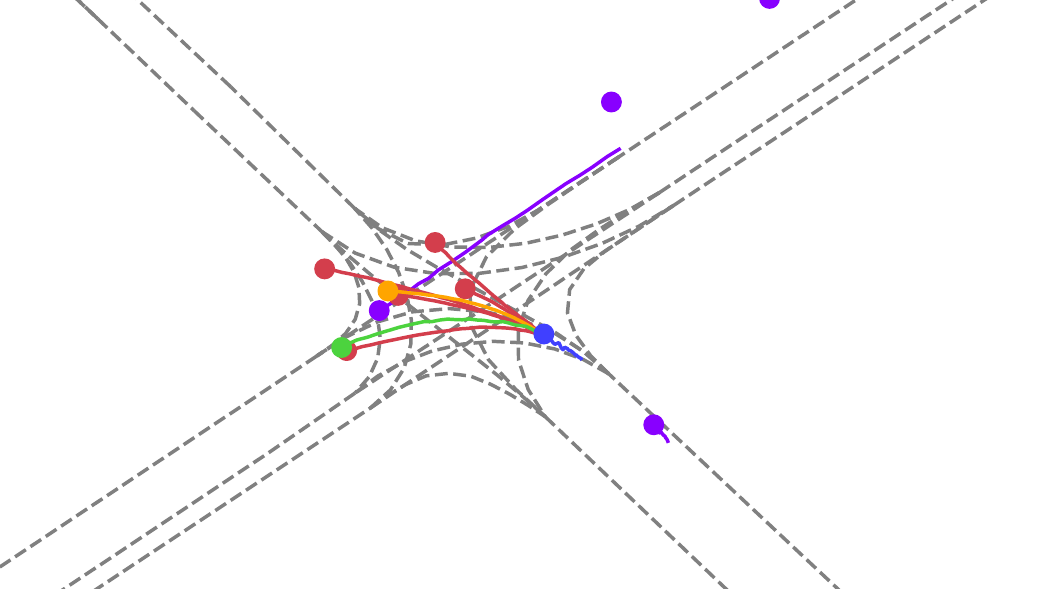}}
	\vspace{-0.1cm}
	\caption{Qualitative results of CRAT-Pred on the Argoverse validation set for three diverse sequences. The past observed trajectory of the target vehicle is colored in blue, the ground-truth future trajectory in green. Predictions are colored in orange and red, with orange corresponding to the most probable future trajectory. Past trajectories of other vehicles are colored in purple. Although not used by the prediction model, road topologies are shown with dashed lines.}
	\label{fig:qualitative_results}
\end{figure*}

Fig. \ref{fig:qualitative_results} shows qualitative results of our proposed prediction model on three diverse sequences of the Argoverse validation set.
The model's capability to make multi-modal predictions can be seen in the low-speed sequence on the right.

\subsection{Self-Attention as a Score for Social Interactions}
Although frequently used for vehicle prediction, the ability of self-attention to learn and describe social interactions has never been analyzed quantitatively until now.
Therefore, we quantitatively analyze if self-attention is able to learn social interactions between vehicles.
We do so, by limiting the vehicles of the original Argoverse dataset in each sequence to a smaller subset, only containing the target vehicle and a maximum of $L_s$ other vehicles.
Two different strategies for vehicle selection are compared:
\begin{enumerate}
	\item \textbf{Euclidean Selection}: Heuristic selection of the $L_s$ closest vehicles to the target vehicle at $t=0$, measured by the Euclidean distance. Euclidean selection with a fixed range is used in some recent approaches \cite{Li2020a_ARXIV, Salzmann2020}.
	\item \textbf{Attention-based Selection}: Selection of the $L_s$ vehicles with the highest attention weights (averaged over all heads) with respect to the target vehicle. During the forward pass of our trained prediction model, the history of each agent and their interactions get encoded. After the last graph convolution layer, the attention weight matrix is calculated and the attention weights of the target vehicle get extracted. The subsequent selection then takes the $L_s$ other vehicles with the highest attention weights.
\end{enumerate}

We claim that an independent trajectory prediction model should achieve a higher performance with the subsets that contain vehicles that are more relevant (here relevance corresponds to the amount of interaction) for the prediction of the target vehicle's trajectory.
Since the attention-based selected subsets directly result from our model, an independent trajectory prediction model must be used in order to evaluate this in an unbiased way.

As an independent model the publicly available state-of-the-art model LaneGCN \cite{Liang2020} is used.
One small adaptation was required.
This adaptation limits the optimization of the model to the trajectory of the target vehicle only, instead of all vehicles jointly.
Since the target vehicle is always located near the center of a sequence, this assures that all of the vehicles that socially interact with the target vehicle are available.

\begin{figure*}[thpb]
	\centering
	\begin{adjustbox}{clip, trim=0cm 0.13cm 0cm 0cm}
		\resizebox{\textwidth}{!}{\input{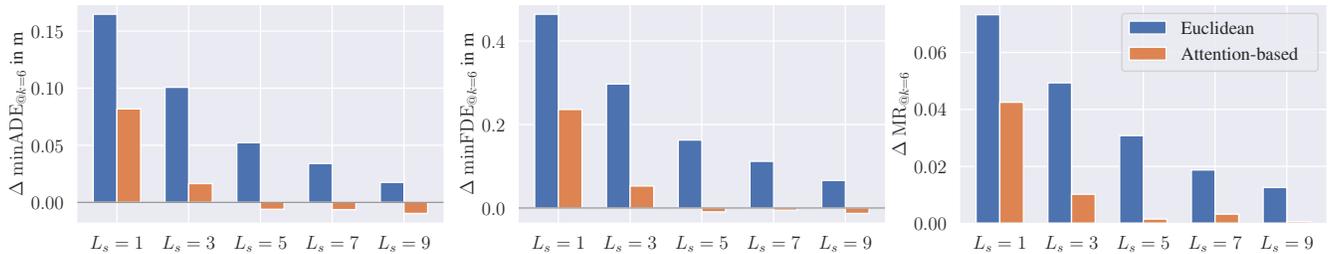}}
	\end{adjustbox}
	\vspace{-0.6cm}
	\caption{Quantitative performance differences of LaneGCN on the Argoverse validation set, trained and validated with the subset selected based on the Euclidean distance (blue) and the attention-based selected subset (orange). Performance is measured in minADE (left), minFDE (center) and MR (right), with the reference value given by LaneGCN trained and validated on the full dataset. Positive performance differences indicate performance degradation and vice versa.}
	\label{fig:attention_selection}
\end{figure*}

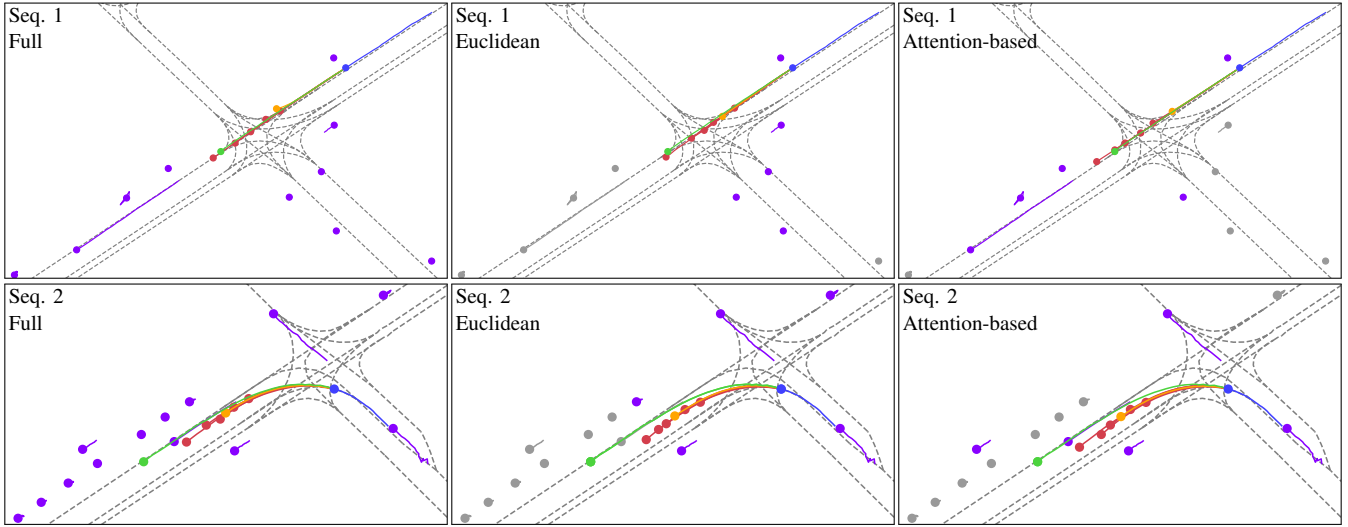
\begin{figure*}[thpb]
	\centering
	\subfloat{%
		\begin{overpic}[width=0.33\textwidth, trim=9.5cm 8.9cm 15.6cm 13.5cm, clip, frame]
			{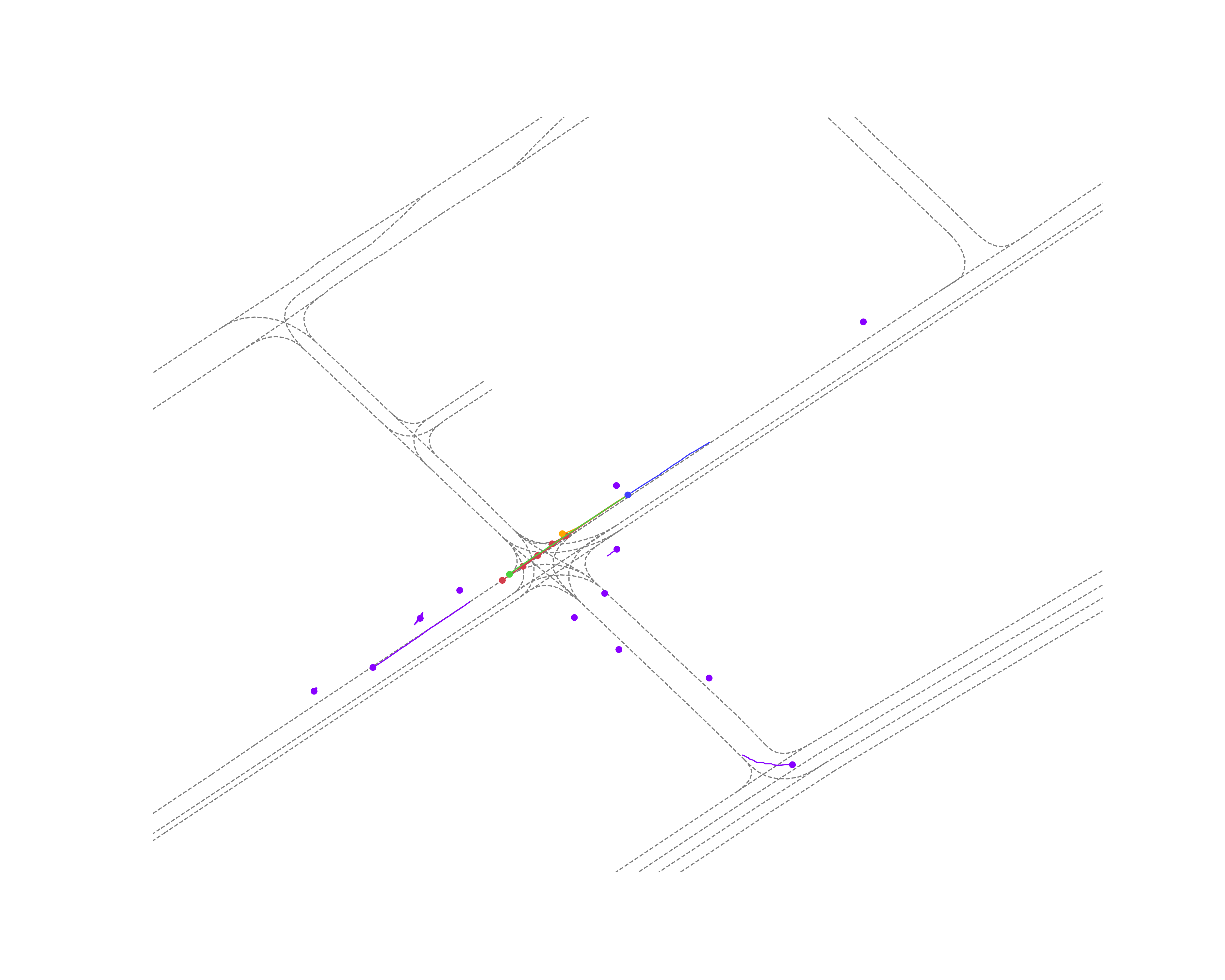}
			\put(1, 58){\footnotesize Seq. 1}
			\put(1, 52){\footnotesize Full}
		\end{overpic}}
	\hfill
	\subfloat{%
		\begin{overpic}[width=0.33\textwidth, trim=9.5cm 8.9cm 15.6cm 13.5cm, clip, frame]
			{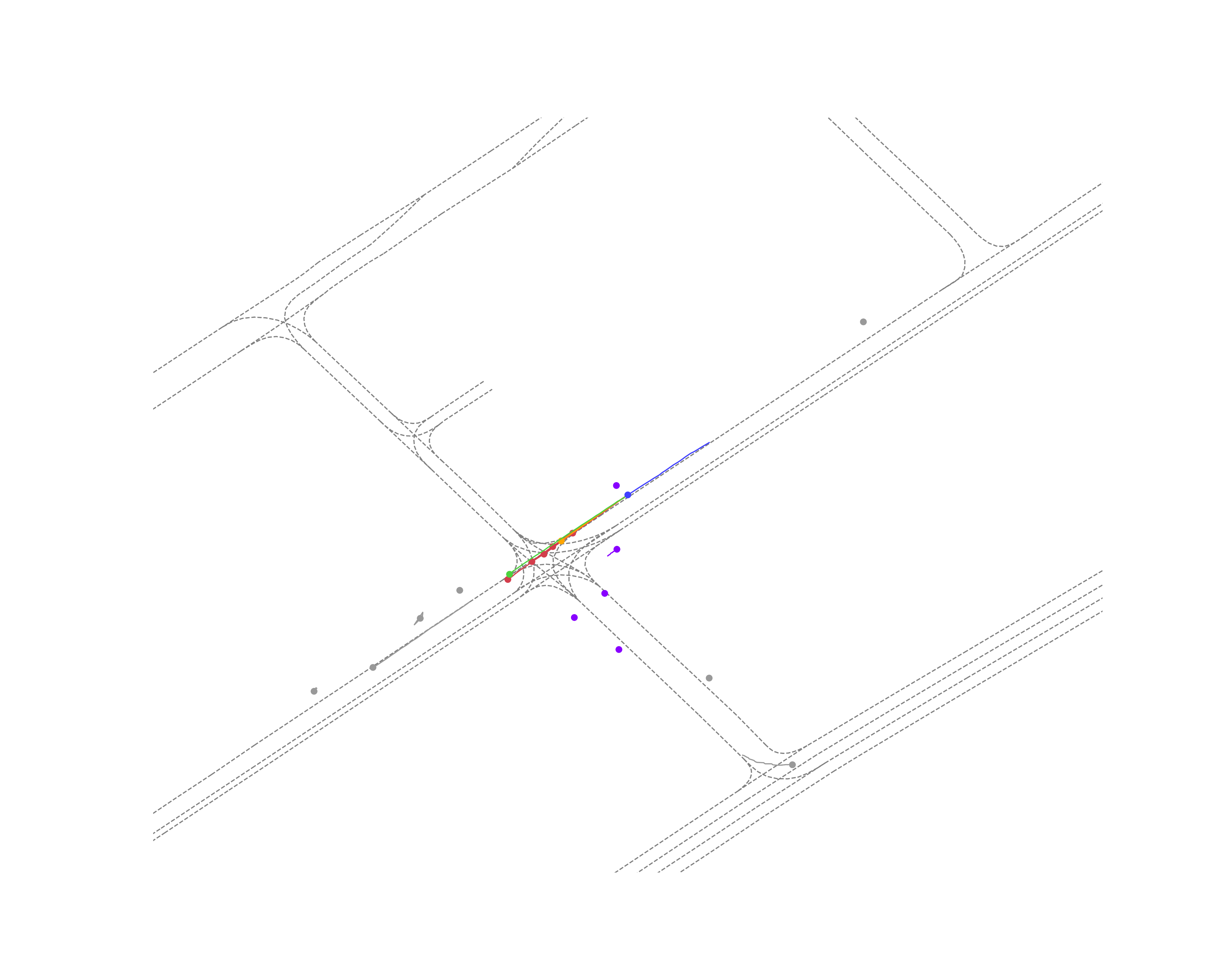}
			\put(1, 58){\footnotesize Seq. 1}
			\put(1, 52){\footnotesize Euclidean}
		\end{overpic}}
	\hfill
	\centering
	\subfloat{%
		\begin{overpic}[width=0.33\textwidth, trim=9.5cm 8.9cm 15.6cm 13.5cm, clip, frame]
			{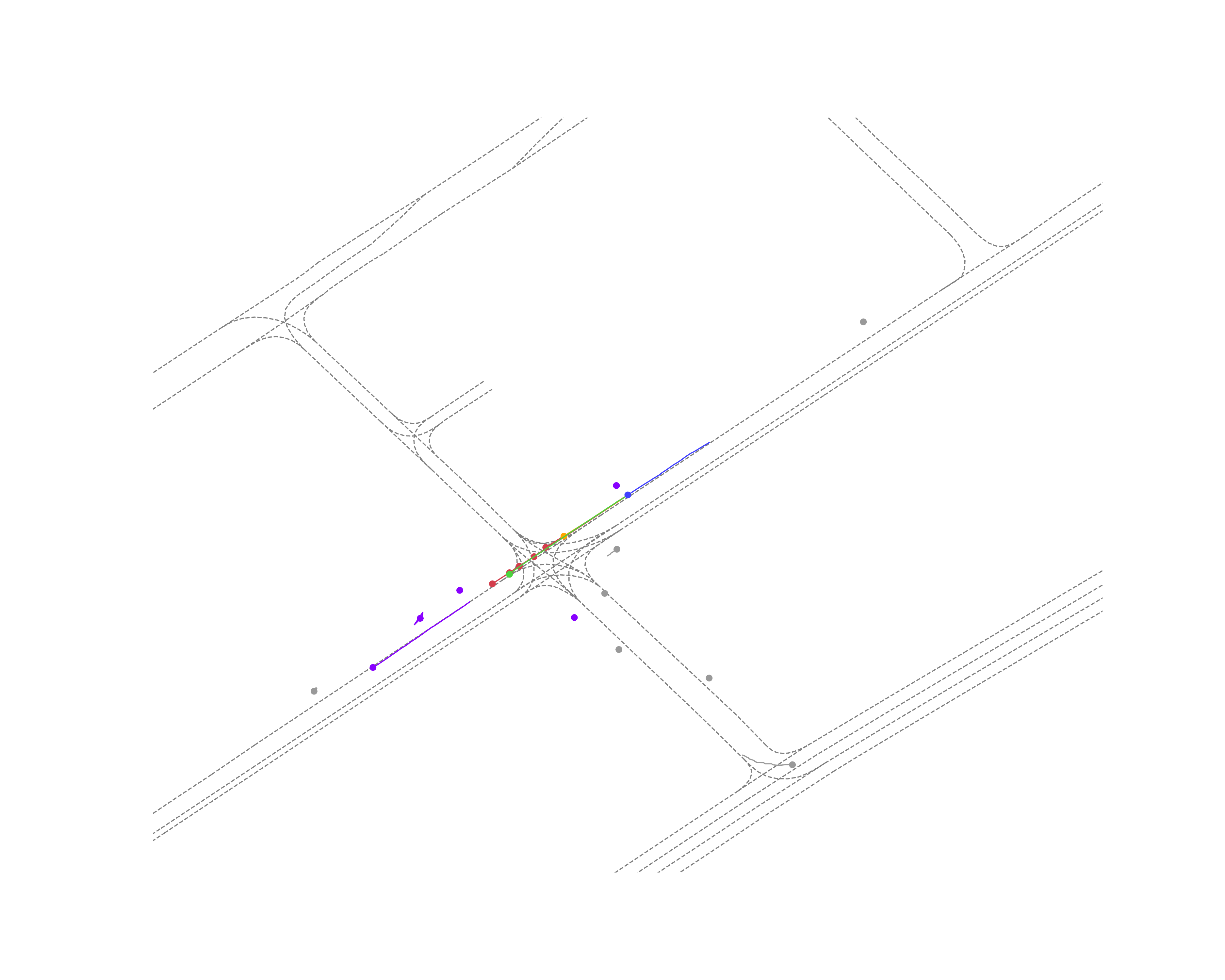}
			\put(1, 58){\footnotesize Seq. 1}
			\put(1, 52){\footnotesize Attention-based}
		\end{overpic}}
	\\[-0.26cm]
	\subfloat{%
		\begin{overpic}[width=0.33\textwidth, trim=12cm 12cm 16cm 13cm, clip, frame]
			{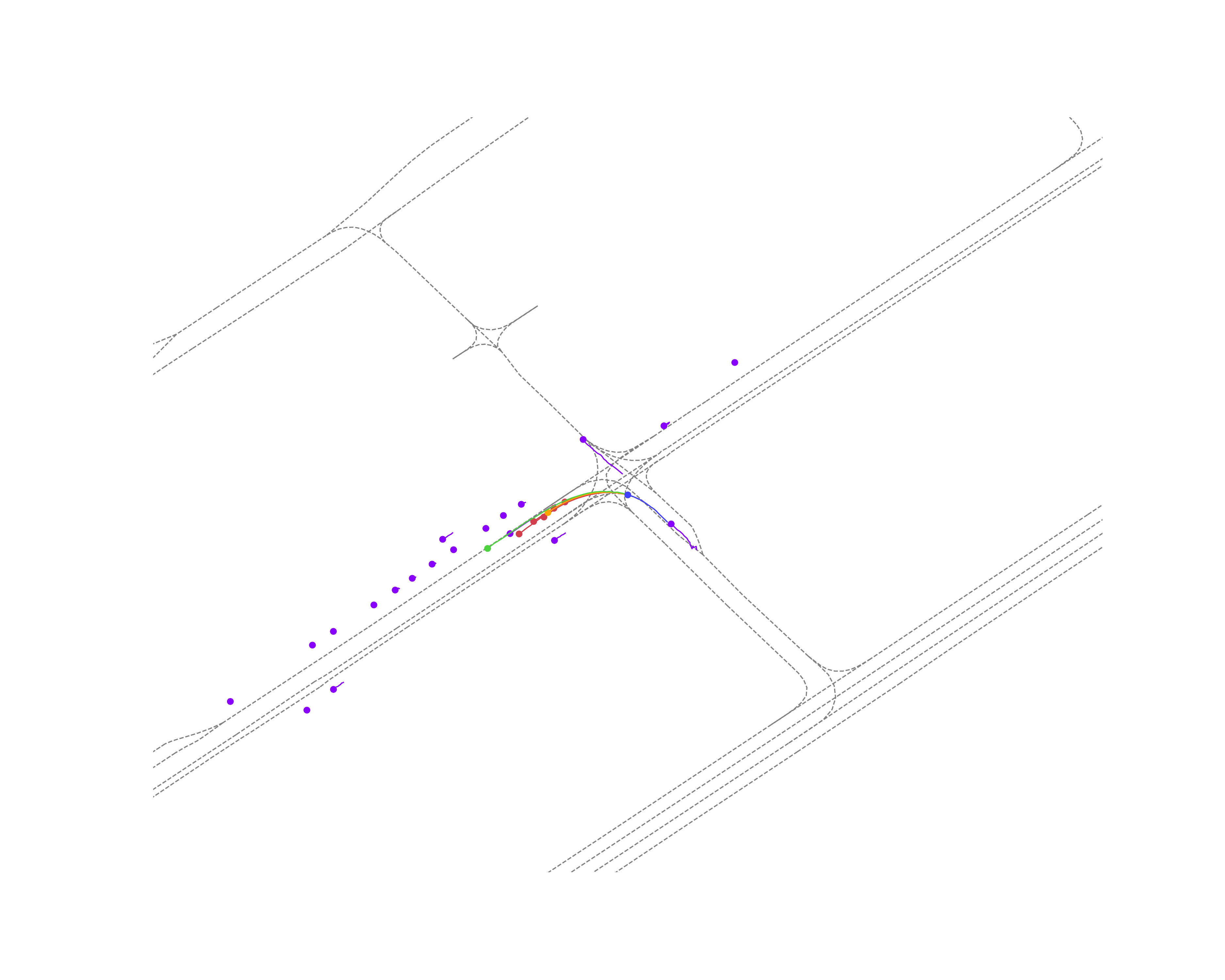}
			\put(1, 50){\footnotesize Seq. 2}
			\put(1, 44){\footnotesize Full}
		\end{overpic}}
	\hfill
	\subfloat{%
		\begin{overpic}[width=0.33\textwidth, trim=12cm 12cm 16cm 13cm, clip, frame]
			{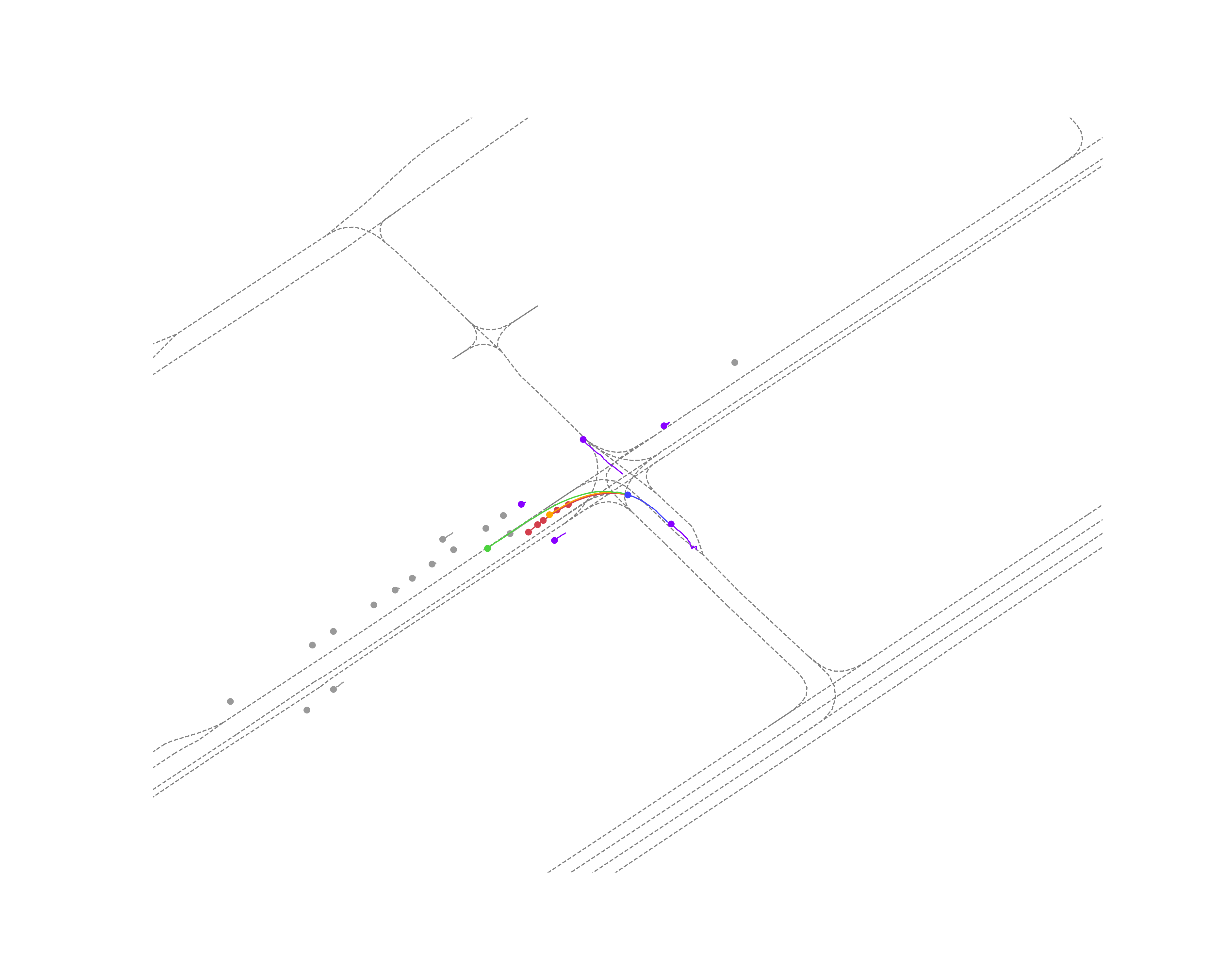}
			\put(1, 50){\footnotesize Seq. 2}
			\put(1, 44){\footnotesize Euclidean}
		\end{overpic}}
	\hfill
	\centering
	\subfloat{%
		\begin{overpic}[width=0.33\textwidth, trim=12cm 12cm 16cm 13cm, clip, frame]
			{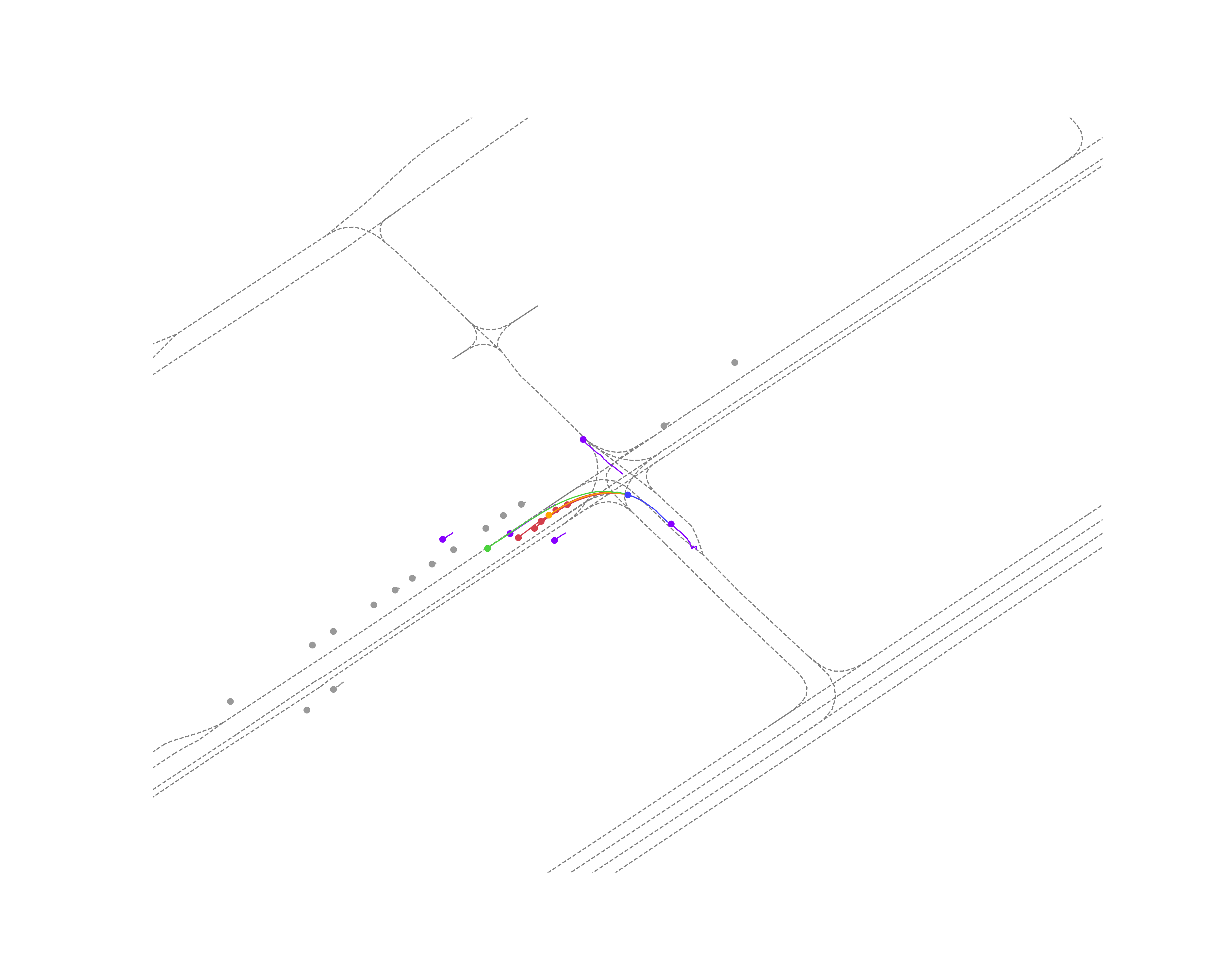}
			\put(1, 50){\footnotesize Seq. 2}
			\put(1, 44){\footnotesize Attention-based}
		\end{overpic}}
	\vspace{-0.1cm}
	\caption{Qualitative results of LaneGCN on the Argoverse validation set, trained and validated with the full dataset (left), the subset selected based on the Euclidean distance (center) and the attention-based selected subset (right). Two sequences are shown and each sequence is limited to a maximum of $L_s = 5$ other vehicles. The past observed trajectory of the target vehicle is colored in blue, the ground-truth future trajectory in green. Predictions are colored in orange and red, with orange corresponding to the most probable future trajectory. Past trajectories of other vehicles, whose selection is done by the different strategies, are colored in purple. Grey circles mark vehicles that are not selected by the corresponding strategy.}
	\label{fig:attention_selection_examples}
\end{figure*}

Absolute performance differences for $L_s \in \{ 1, 3, 5, 7, 9 \}$ are plotted in Fig. \ref{fig:attention_selection}.
The reference value is set by LaneGCN trained and validated on the full dataset, which is averaging more than $15$ vehicles in each sequence, resulting in minADE$_{@k=6} = 1.398$, minFDE$_{@k=6} = 3.051$ and MR$_{@k=6} = 0.505$.
It is observable that the attention-based selection leads to a superior performance for all cases, when comparing it to the Euclidean selection.
Interestingly, for $L_s \in \{ 5, 7, 9 \}$ LaneGCN trained and validated on the attention-based selected subset results in a lower minADE$_{@k=6}$ and minFDE$_{@k=6}$ than LaneGCN trained and validated on the whole dataset.

While prior publications in the field of trajectory prediction applied self-attention and observed an improvement in the performance of a prediction model, our vehicle selection experiment quantitatively confirms what was mainly assumed before: Self-attention is indeed able to learn social interaction between vehicles, with the weights representing a measurable interaction score.

Fig. \ref{fig:attention_selection_examples} qualitatively shows the available vehicle data and the resulting trajectory predictions on two exemplary sequences for $L_s = 5$.
In Sequence 1, the attention-based selection focuses mainly on vehicles driving in front of the target vehicle.
This also reflects the intuitive behavior of a human driver.
Even more crucially, in Sequence 2, the Euclidean selection misses out on a vehicle traveling with the same trajectory as the target vehicle in the future.
The attention-based selection includes this vehicle, which leads to better prediction results.

\section{Conclusion}
This paper proposes a simple yet effective trajectory prediction model for vehicles, achieving state-of-the-art performance without using map information.
In contrast to approaches that offer a comparable performance, it requires significantly less model parameters.
The model uses an LSTM for the temporal encoding of vehicle features and then applies a GNN and multi-head self-attention to model social interactions.
A simple linear residual layer is then used to generate the trajectory predictions.

While multi-head self-attention does boost the model's performance, the results of the vehicle selection experiment indicate another key property of the self-attention mechanism:
It is able to learn social interactions and therefore rate interactions between vehicles.

\addtolength{\textheight}{-4.8cm}





\section*{Acknowledgment}
This publication was created as part of the research project "KI Delta Learning" (project number: 19A19013A) funded by the Federal Ministry for Economic Affairs and Energy (BMWi) on the basis of a decision by the German Bundestag.

\clearpage
\bibliographystyle{IEEEtran}
\bibliography{Literature}

\begin{thebibliography}{10}
\providecommand{\url}[1]{#1}
\csname url@samestyle\endcsname
\providecommand{\newblock}{\relax}
\providecommand{\bibinfo}[2]{#2}
\providecommand{\BIBentrySTDinterwordspacing}{\spaceskip=0pt\relax}
\providecommand{\BIBentryALTinterwordstretchfactor}{4}
\providecommand{\BIBentryALTinterwordspacing}{\spaceskip=\fontdimen2\font plus
\BIBentryALTinterwordstretchfactor\fontdimen3\font minus
  \fontdimen4\font\relax}
\providecommand{\BIBforeignlanguage}[2]{{%
\expandafter\ifx\csname l@#1\endcsname\relax
\typeout{** WARNING: IEEEtran.bst: No hyphenation pattern has been}%
\typeout{** loaded for the language `#1'. Using the pattern for}%
\typeout{** the default language instead.}%
\else
\language=\csname l@#1\endcsname
\fi
#2}}
\providecommand{\BIBdecl}{\relax}
\BIBdecl

\bibitem{Hong2019}
J.~{Hong}, B.~{Sapp}, and J.~{Philbin}, ``Rules of the road: Predicting driving
  behavior with a convolutional model of semantic interactions,'' in \emph{2019
  IEEE/CVF Conference on Computer Vision and Pattern Recognition (CVPR)}, 2019,
  pp. 8446--8454.

\bibitem{Li2020a_ARXIV}
X.~Li, X.~Ying, and M.~C. Chuah, ``Grip++: Enhanced graph-based
  interaction-aware trajectory prediction for autonomous driving,'' 2020.

\bibitem{Liang2020}
M.~Liang, B.~Yang, R.~Hu, Y.~Chen, R.~Liao, S.~Feng, and R.~Urtasun, ``Learning
  lane graph representations for motion forecasting,'' in \emph{Computer Vision
  -- ECCV 2020}, A.~Vedaldi, H.~Bischof, T.~Brox, and J.-M. Frahm, Eds.\hskip
  1em plus 0.5em minus 0.4em\relax Cham: Springer International Publishing,
  2020, pp. 541--556.

\bibitem{Schoeller2020}
C.~Sch{\"o}ller, V.~Aravantinos, F.~Lay, and A.~Knoll, ``What the constant
  velocity model can teach us about pedestrian motion prediction,'' \emph{IEEE
  Robotics and Automation Letters}, vol.~5, no.~2, pp. 1696--1703, 2020.

\bibitem{Xie2018}
T.~Xie and J.~C. Grossman, ``Crystal graph convolutional neural networks for an
  accurate and interpretable prediction of material properties,'' \emph{Phys.
  Rev. Lett.}, vol. 120, p. 145301, Apr 2018.

\bibitem{Chang2019}
M.-F. Chang, J.~Lambert, P.~Sangkloy, J.~Singh, S.~Bak, A.~Hartnett, D.~Wang,
  P.~Carr, S.~Lucey, D.~Ramanan, and J.~Hays, ``Argoverse: 3d tracking and
  forecasting with rich maps,'' in \emph{2019 IEEE/CVF Conference on Computer
  Vision and Pattern Recognition (CVPR)}, 2019, pp. 8740--8749.

\bibitem{Deo2018}
N.~Deo and M.~M. Trivedi, ``Convolutional social pooling for vehicle trajectory
  prediction,'' in \emph{2018 IEEE/CVF Conference on Computer Vision and
  Pattern Recognition Workshops (CVPRW)}, 2018, pp. 1581--1589.

\bibitem{Alahi2016}
A.~{Alahi}, K.~{Goel}, V.~{Ramanathan}, A.~{Robicquet}, L.~{Fei-Fei}, and
  S.~{Savarese}, ``Social lstm: Human trajectory prediction in crowded
  spaces,'' in \emph{2016 IEEE Conference on Computer Vision and Pattern
  Recognition (CVPR)}, 2016, pp. 961--971.

\bibitem{Chai2020}
Y.~Chai, B.~Sapp, M.~Bansal, and D.~Anguelov, ``Multipath: Multiple
  probabilistic anchor trajectory hypotheses for behavior prediction,'' in
  \emph{Proceedings of the Conference on Robot Learning}, ser. Proceedings of
  Machine Learning Research, vol. 100.\hskip 1em plus 0.5em minus 0.4em\relax
  PMLR, 2020, pp. 86--99.

\bibitem{Djuric2020}
N.~Djuric, V.~Radosavljevic, H.~Cui, T.~Nguyen, F.-C. Chou, T.-H. Lin,
  N.~Singh, and J.~Schneider, ``Uncertainty-aware short-term motion prediction
  of traffic actors for autonomous driving,'' in \emph{2020 IEEE Winter
  Conference on Applications of Computer Vision (WACV)}, 2020, pp. 2084--2093.

\bibitem{Strohbeck2020}
J.~Strohbeck, V.~Belagiannis, J.~M{\"u}ller, M.~Schreiber, M.~Herrmann,
  D.~Wolf, and M.~Buchholz, ``Multiple trajectory prediction with deep temporal
  and spatial convolutional neural networks,'' in \emph{2020 IEEE/RSJ
  International Conference on Intelligent Robots and Systems (IROS)}, 2020, pp.
  1992--1998.

\bibitem{Casas2020}
S.~Casas, C.~Gulino, S.~Suo, K.~Luo, R.~Liao, and R.~Urtasun, ``Implicit latent
  variable model for scene-consistent motion forecasting,'' in \emph{Computer
  Vision -- ECCV 2020}, A.~Vedaldi, H.~Bischof, T.~Brox, and J.-M. Frahm,
  Eds.\hskip 1em plus 0.5em minus 0.4em\relax Cham: Springer International
  Publishing, 2020, pp. 624--641.

\bibitem{Salzmann2020}
T.~Salzmann, B.~Ivanovic, P.~Chakravarty, and M.~Pavone, ``Trajectron++:
  Dynamically-feasible trajectory forecasting with heterogeneous data,'' in
  \emph{Computer Vision -- ECCV 2020}, A.~Vedaldi, H.~Bischof, T.~Brox, and
  J.-M. Frahm, Eds.\hskip 1em plus 0.5em minus 0.4em\relax Cham: Springer
  International Publishing, 2020, pp. 683--700.

\bibitem{Gilmer2017}
J.~Gilmer, S.~S. Schoenholz, P.~F. Riley, O.~Vinyals, and G.~E. Dahl, ``Neural
  message passing for quantum chemistry,'' in \emph{Proceedings of the 34th
  International Conference on Machine Learning - Volume 70}, ser.
  ICML'17.\hskip 1em plus 0.5em minus 0.4em\relax JMLR.org, 2017, pp.
  1263--1272.

\bibitem{Tang2019}
C.~Tang and R.~R. Salakhutdinov, ``Multiple futures prediction,'' in
  \emph{Advances in Neural Information Processing Systems}, H.~Wallach,
  H.~Larochelle, A.~Beygelzimer, F.~d\textquotesingle Alch\'{e}-Buc, E.~Fox,
  and R.~Garnett, Eds., vol.~32.\hskip 1em plus 0.5em minus 0.4em\relax Curran
  Associates, Inc., 2019.

\bibitem{Messaoud2021}
K.~Messaoud, I.~Yahiaoui, A.~Verroust-Blondet, and F.~Nashashibi, ``Attention
  based vehicle trajectory prediction,'' \emph{IEEE Transactions on Intelligent
  Vehicles}, vol.~6, no.~1, pp. 175--185, 2021.

\bibitem{Mercat2020}
J.~{Mercat}, T.~{Gilles}, N.~{El Zoghby}, G.~{Sandou}, D.~{Beauvois}, and G.~P.
  {Gil}, ``Multi-head attention for multi-modal joint vehicle motion
  forecasting,'' in \emph{2020 IEEE International Conference on Robotics and
  Automation (ICRA)}, 2020, pp. 9638--9644.

\bibitem{Gao2020}
J.~{Gao}, C.~{Sun}, H.~{Zhao}, Y.~{Shen}, D.~{Anguelov}, C.~{Li}, and
  C.~{Schmid}, ``Vectornet: Encoding hd maps and agent dynamics from vectorized
  representation,'' in \emph{2020 IEEE/CVF Conference on Computer Vision and
  Pattern Recognition (CVPR)}, 2020, pp. 11\,522--11\,530.

\bibitem{Vaswani2017}
A.~Vaswani, N.~Shazeer, N.~Parmar, J.~Uszkoreit, L.~Jones, A.~N. Gomez, L.~u.
  Kaiser, and I.~Polosukhin, ``Attention is all you need,'' in \emph{Advances
  in Neural Information Processing Systems}, I.~Guyon, U.~V. Luxburg,
  S.~Bengio, H.~Wallach, R.~Fergus, S.~Vishwanathan, and R.~Garnett, Eds.,
  vol.~30.\hskip 1em plus 0.5em minus 0.4em\relax Curran Associates, Inc.,
  2017.

\bibitem{Khandelwal2020_ARXIV}
S.~Khandelwal, W.~Qi, J.~Singh, A.~Hartnett, and D.~Ramanan, ``What-if motion
  prediction for autonomous driving,'' 2020.

\bibitem{Ma2019}
Y.~Ma, X.~Zhu, S.~Zhang, R.~Yang, W.~Wang, and D.~Manocha, ``Trafficpredict:
  Trajectory prediction for heterogeneous traffic-agents,'' in \emph{The
  Thirty-Third {AAAI} Conference on Artificial Intelligence, {AAAI}
  2019}.\hskip 1em plus 0.5em minus 0.4em\relax {AAAI} Press, 2019, pp.
  6120--6127.

\bibitem{Li2020b}
J.~Li, H.~Ma, Z.~Zhang, and M.~Tomizuka, ``Social-wagdat: Interaction-aware
  trajectory prediction via wasserstein graph double-attention network,'' 2020.

\bibitem{Kim2021_ARXIV}
B.~Kim, S.~H. Park, S.~Lee, E.~Khoshimjonov, D.~Kum, J.~Kim, J.~S. Kim, and
  J.~W. Choi, ``Lapred: Lane-aware prediction of multi-modal future
  trajectories of dynamic agents,'' 2021.

\bibitem{Zeng2021_ARXIV}
W.~Zeng, M.~Liang, R.~Liao, and R.~Urtasun, ``Lanercnn: Distributed
  representations for graph-centric motion forecasting,'' 2021.

\bibitem{Ye2021}
M.~Ye, T.~Cao, and Q.~Chen, ``Tpcn: Temporal point cloud networks for motion
  forecasting,'' in \emph{2021 IEEE/CVF Conference on Computer Vision and
  Pattern Recognition (CVPR)}, 2021, pp. 11\,318--11\,327.

\bibitem{Liu2021}
Y.~Liu, J.~Zhang, L.~Fang, Q.~Jiang, and B.~Zhou, ``Multimodal motion
  prediction with stacked transformers,'' in \emph{Proceedings of the IEEE/CVF
  Conference on Computer Vision and Pattern Recognition (CVPR)}, June 2021, pp.
  7577--7586.

\bibitem{Guzmanrivera2012}
A.~Guzm\'{a}n-rivera, D.~Batra, and P.~Kohli, ``Multiple choice learning:
  Learning to produce multiple structured outputs,'' in \emph{Advances in
  Neural Information Processing Systems}, F.~Pereira, C.~J.~C. Burges,
  L.~Bottou, and K.~Q. Weinberger, Eds., vol.~25.\hskip 1em plus 0.5em minus
  0.4em\relax Curran Associates, Inc., 2012.

\bibitem{Kingma2015}
D.~P. Kingma and J.~Ba, ``Adam: {A} method for stochastic optimization,'' in
  \emph{3rd International Conference on Learning Representations, {ICLR} 2015},
  Y.~Bengio and Y.~LeCun, Eds., 2015.

\bibitem{Paszke2019}
A.~Paszke, S.~Gross, F.~Massa, A.~Lerer, J.~Bradbury, G.~Chanan, T.~Killeen,
  Z.~Lin, N.~Gimelshein, L.~Antiga, A.~Desmaison, A.~Kopf, E.~Yang, Z.~DeVito,
  M.~Raison, A.~Tejani, S.~Chilamkurthy, B.~Steiner, L.~Fang, J.~Bai, and
  S.~Chintala, ``Pytorch: An imperative style, high-performance deep learning
  library,'' in \emph{Advances in Neural Information Processing Systems 32},
  H.~Wallach, H.~Larochelle, A.~Beygelzimer, F.~d\textquotesingle
  Alch\'{e}-Buc, E.~Fox, and R.~Garnett, Eds.\hskip 1em plus 0.5em minus
  0.4em\relax Curran Associates, Inc., 2019, pp. 8024--8035.

\bibitem{Fey2019}
M.~Fey and J.~E. Lenssen, ``Fast graph representation learning with {PyTorch
  Geometric},'' in \emph{ICLR Workshop on Representation Learning on Graphs and
  Manifolds}, 2019.

\end{thebibliography}

\end{document}